%% file: main.tex
\documentclass[conference]{IEEEtran}

\usepackage{cite}
\usepackage{amsmath,amssymb,amsfonts}
\usepackage{algorithm}
\usepackage{algpseudocode}
\usepackage{graphicx}
\usepackage{textcomp}
\usepackage{xcolor}
\usepackage{booktabs}
\usepackage[hidelinks]{hyperref}
\usepackage{tikz}
\usepackage{pgfplots}
\pgfplotsset{compat=1.18}

\def\BibTeX{{\rm B\kern-.05em{\sc i\kern-.025em b}\kern-.08em
    T\kern-.1667em\lower.7ex\hbox{E}\kern-.125emX}}

\IEEEoverridecommandlockouts
\title{Reason--Imagine--Act: Closed-Loop LLM Decision Making with World Models for Autonomous Driving}

\author{
\IEEEauthorblockN{Zhengqi Sun\textsuperscript{1},
Yiwen Sun\textsuperscript{2,3,*},
Boxuan Liu\textsuperscript{4},
Tailai Chen\textsuperscript{5},
Tianxu Guo\textsuperscript{6},
and Jiabin Liu\textsuperscript{6}}
\IEEEauthorblockA{\textsuperscript{1}Department of Information Management, Peking University, Beijing 100871, China}
\IEEEauthorblockA{\textsuperscript{2}School of Intelligence Science and Technology, Peking University, Beijing 100871, China}
\IEEEauthorblockA{\textsuperscript{3}State Key Laboratory of General Artificial Intelligence, BIGAI, Beijing 100080, China}
\IEEEauthorblockA{\textsuperscript{4}Yuanpei College, Peking University, Beijing 100871, China}
\IEEEauthorblockA{\textsuperscript{5}China Agricultural University, Beijing, China}
\IEEEauthorblockA{\textsuperscript{6}CRSC Research \& Design Institute Group Co., Ltd., Beijing, China}
\IEEEauthorblockA{\textsuperscript{*}Corresponding author: Yiwen Sun (sunyiwen@pku.edu.cn)}
\thanks{This work was supported by the National Natural Science Foundation of China under Grant No. 62503015.}
}

\begin{document}
\maketitle
\pagestyle{empty}
\thispagestyle{empty}

\begin{abstract}
Large language models (LLMs) are promising for autonomous driving, but semantics-only decision policies can yield physically unsafe behavior in dynamic traffic. Existing methods either perform online language reasoning without explicit dynamics verification or use world models mainly in offline pipelines, leaving a gap between semantic intent and physical feasibility at decision time. We propose \textbf{Reason--Imagine--Act (RIA)}, a closed-loop framework that couples an LLM reasoner with an action-conditioned world model for online safety verification. At each step, the LLM proposes an action template and candidate sub-actions, the world model performs short-horizon rollouts, and a safety scorer selects the safest executable action with feedback to the next reasoning step. Under a unified CARLA point-goal protocol (1000 episodes), RIA achieves 80.05\% route completion, 51.10\% arrival rate, and 0.20\% collision rate. Under the same closed-loop interface, RIA consistently outperforms training-free baselines, including CARLA TM and MADA, on core closed-loop metrics. For reproducibility, code is available at \url{https://github.com/pku-smart-city/source_code/tree/main/RIA}.
\end{abstract}

\begin{IEEEkeywords}
autonomous driving, large language models, world models, closed-loop decision making, safety verification
\end{IEEEkeywords}

\input{sections/intro.tex}

\input{sections/related_work.tex}

\input{sections/method.tex}

\input{sections/experiments.tex}

\input{sections/conclusion.tex}

\IEEEtriggeratref{15}
\bibliographystyle{IEEEtran}
\bibliography{ref}

\end{document}

%% file: sections/intro.tex
\section{Introduction}
Autonomous driving is undergoing a profound paradigm shift from traditional modular pipelines---typically decomposed into perception, prediction, and planning---towards model-centric decision-making driven by large foundation models. Such large-scale pre-trained models have demonstrated strong transferability and generalization, motivating the community to revisit how ``intelligence'' should be implemented in the driving stack. In particular, Large Language Models (LLMs) introduce powerful common-sense reasoning and semantic understanding, enabling agents to interpret traffic rules, explain decisions, and generalize to previously unseen long-tail situations.

\begin{figure}[t]
    \centering
    \includegraphics[width=0.96\linewidth,trim=16 10 16 10,clip]{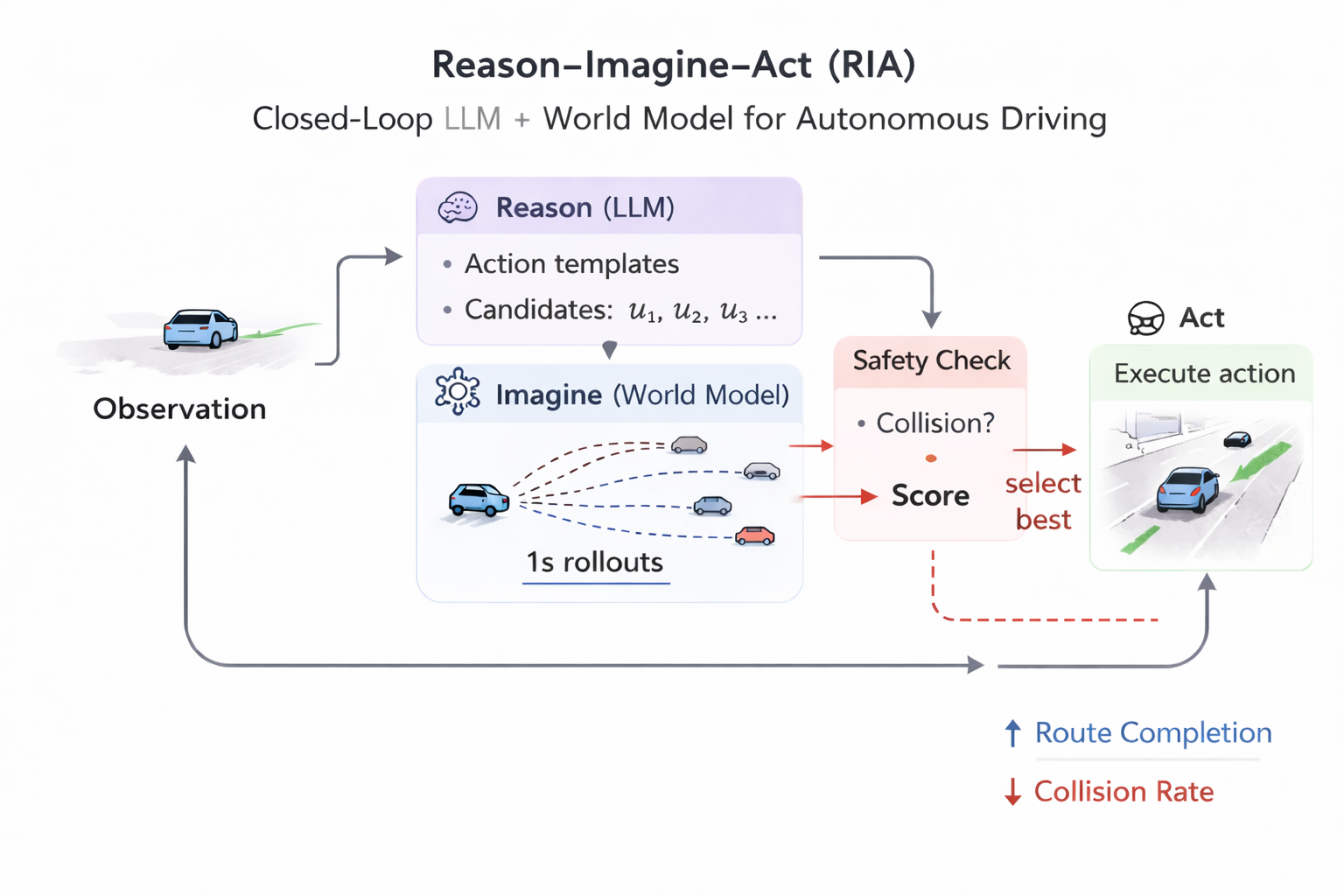}
    \caption{Overview of the proposed Reason--Imagine--Act framework. The LLM proposes high-level intents and candidates, the world model imagines short-horizon futures, and the safety scorer selects the safest executable action with feedback to close the loop.}
    \label{fig:intro_ria}
\end{figure}

Despite these promising capabilities, deploying LLMs directly for real-time driving decisions faces a fundamental obstacle: \emph{the lack of physical grounding}. Trained primarily on textual or multimodal semantic corpora, LLMs often behave as a ``brain in a vat'': they can generate logically coherent high-level commands (e.g., ``execute emergency avoidance''), yet they do not precisely capture vehicle dynamic constraints, nor can they reliably anticipate the numerical consequences of commands in continuous physical space. This decoupling between semantic logic and physical reality leads to severe \emph{physical hallucinations}, where the generated intention is kinematically/dynamically infeasible and thus threatens driving safety.

Meanwhile, World Models (WMs) have shown substantial potential for capturing physical regularities by learning environment evolution and supporting multi-step prediction. However, existing studies on world models for autonomous driving predominantly emphasize offline video generation quality or their use as auxiliary predictors during pre-training, while tight integration with LLM-level reasoning for \emph{real-time} decision-making is still limited. As a result, high-level reasoning is often ``rule-aware but physics-agnostic'', whereas world models are ``dynamics-aware but intent-agnostic'', leaving a clear gap between semantic decision logic and physically grounded execution.

To bridge this gap, we propose a novel closed-loop autonomous driving framework, termed \textbf{Reason--Imagine--Act}. The key idea is to tightly couple LLM-based high-level semantic reasoning with WM-based numerical dynamics prediction in the decision loop. Concretely, the LLM first generates a high-level driving intent; the world model then performs multi-step physical rollouts conditioned on this intent to reveal potential collision risks or kinematic/dynamic violations \emph{before} execution. The resulting physical feedback further refines action selection and is fed back to the reasoning module, enabling a ``think-before-acting'' loop that mitigates physically unsafe decisions.

We validate the proposed framework in the high-fidelity 3D simulator CARLA under point-goal navigation in complex urban dynamics. Experimental results show that, compared with a vanilla LLM baseline, our framework significantly improves route completion while substantially reducing the Collision Rate (ColR). Under an identical closed-loop interface, our method also outperforms other training-free baselines in challenging scenarios, demonstrating the practical value of integrating semantic reasoning with physically grounded imagination.

In this paper, we focus on \emph{inference-time} deployment without parameter updates (i.e., no fine-tuning or additional policy training). Accordingly, our comparisons are restricted to methods that can be reproduced under the same CARLA closed-loop interface and evaluation protocol.

\noindent\textbf{Contributions.} Our main contributions are three-fold:
\begin{itemize}
    \item To the best of our knowledge, this is the first work to tightly couple an LLM reasoner with an action-conditioned world model in a closed-loop driving decision pipeline for online \emph{think-before-acting} verification.
    \item We propose \textbf{Reason--Imagine--Act}, a WM-augmented decision mechanism that evaluates template-conditioned candidate sub-actions via short-horizon rollouts and feeds physical verification signals back to the LLM for iterative correction.
    \item Under the same closed-loop interface, our method consistently outperforms comparable training-free baselines, achieving stronger route-completion and collision outcomes. In closed-loop CARLA evaluation, compared with the LLM w/o WM verification variant, our full method improves RC from 61.21\% to 80.05\% and AR from 30.20\% to 51.10\%, while reducing ColR from 0.40\% to 0.20\%.
\end{itemize}

%% file: sections/related_work.tex
\section{Related Work}

\subsection{Large Language Models for Autonomous Driving}
With the rapid progress of large foundation models, recent research has explored leveraging
large language/vision-language models to improve decision making, interpretability, and
generalization in autonomous driving \cite{li2024llmsurvey}.
A first line of work treats LLMs (or multimodal LLMs) as \emph{high-level reasoners} that operate on
structured scene descriptions or learned tokens, producing semantic decisions or plans that can be
executed by downstream controllers or planners \cite{wendilu,tiandrivevlm}.
These systems emphasize commonsense reasoning, rule understanding, and explanatory outputs,
and often incorporate memory, reflection, or hierarchical planning to handle long-horizon tasks.
Related ITSC studies also investigate real-vehicle personalization and human-language interaction,
including field-tested command-to-control systems with memory adaptation \cite{cui2024personalized}
and multimodal warning generation for driver-assistance personalization \cite{xu2024llmwarning}.

A second line of work pushes toward \emph{end-to-end closed-loop} driving with LLMs/VLMs by
conditioning on multi-view sensors (and sometimes vehicle states) and directly generating trajectories
or low-level control signals \cite{shao2024lmdrive,xu2024drivegpt4,xu2025drivegpt4}.
To better balance reasoning quality and real-time constraints, recent approaches introduce
efficient inference schemes such as switching between ``fast'' and ``slow'' modes or progressively
refining trajectories with language-vision synergy \cite{zhouautovla,chen2025solve}.
Despite encouraging progress, many LLM-based driving systems still struggle with \emph{grounding}
under distribution shift: high-level decisions that are semantically plausible often become physically
unsafe when facing complex urban dynamics, rare interactions, or out-of-distribution motion patterns.
This motivates mechanisms that can enforce feasibility and safety through explicit physical feedback
during inference, rather than relying solely on semantic consistency.

\subsection{World Models for Autonomous Driving}
World models aim to learn predictive representations of environment evolution, enabling
counterfactual reasoning, planning, and data augmentation.
Early successes in model-based RL and latent dynamics modeling motivate adopting learned
forward models for decision making in driving-like domains \cite{li2024think2drive,wang2025adawm}.
In autonomous driving, a prominent direction focuses on \emph{generative world models} that synthesize
future driving observations (e.g., multi-view videos) for simulation, augmentation, or evaluation
\cite{wang2024drivedreamer,zhao2025drivedreamer}.
These models are valuable for controllable scenario generation and improving robustness of perception and prediction,
but are often used offline.

More recent efforts emphasize \emph{structured 3D/BEV world representations} for forecasting and planning.
Occupancy-centric world models predict the evolution of fine-grained 3D scene states and can support planning
by evaluating candidate ego actions in the predicted future \cite{zheng2024occworld,yang2025driving}.
Related works also explore feeding compact 3D representations into large models (e.g., occupancy tokens) to improve
spatial understanding and forecasting \cite{xu2025occ}.
However, integrating world models into real-time decision loops remains challenging: many systems
either (i) use world models primarily as offline generators or pretraining auxiliaries, or
(ii) perform planning with limited interaction between high-level reasoning and low-level dynamics.

In contrast to using world models solely for generation or as passive predictors, our work highlights
the importance of \emph{closed-loop coupling} between high-level semantic reasoning and action-conditioned
dynamics rollouts, so that candidate intents can be assessed and refined with numerical physical feedback
before execution.

%% file: sections/method.tex
\section{Methodology}
\label{sec:method}

\subsection{Overview of Reason--Imagine--Act}
\label{sec:ria_overview}
We propose a closed-loop decision-making framework, \textbf{Reason--Imagine--Act (RIA)}, to align
high-level semantic reasoning with low-level physical feasibility in autonomous driving.
At each control cycle $t$, RIA couples a large language model (LLM) with an action-conditioned world model (WM)
to perform \emph{think-before-acting} verification.
Here $t$ denotes the discrete control-step index.

\textbf{High-level idea.}
\textbf{Reason:} the LLM ingests the current state and environment description together with short-term behavior memory,
and selects a \emph{discrete} driving \emph{action template} (e.g., lane change, speed adjustment).
\textbf{Imagine:} conditioned on the chosen template, the WM performs short-horizon rollouts over a small set of
\emph{sub-actions} (parameterizations / primitives) to predict future trajectories and interactions.
\textbf{Act:} the system selects the safest sub-action by a safety score and executes it through a TM-based control backend,
while returning the verification outcomes and execution summary back to the LLM to close the loop.

\begin{figure*}[!t]
    \centering
    \includegraphics[width=\textwidth]{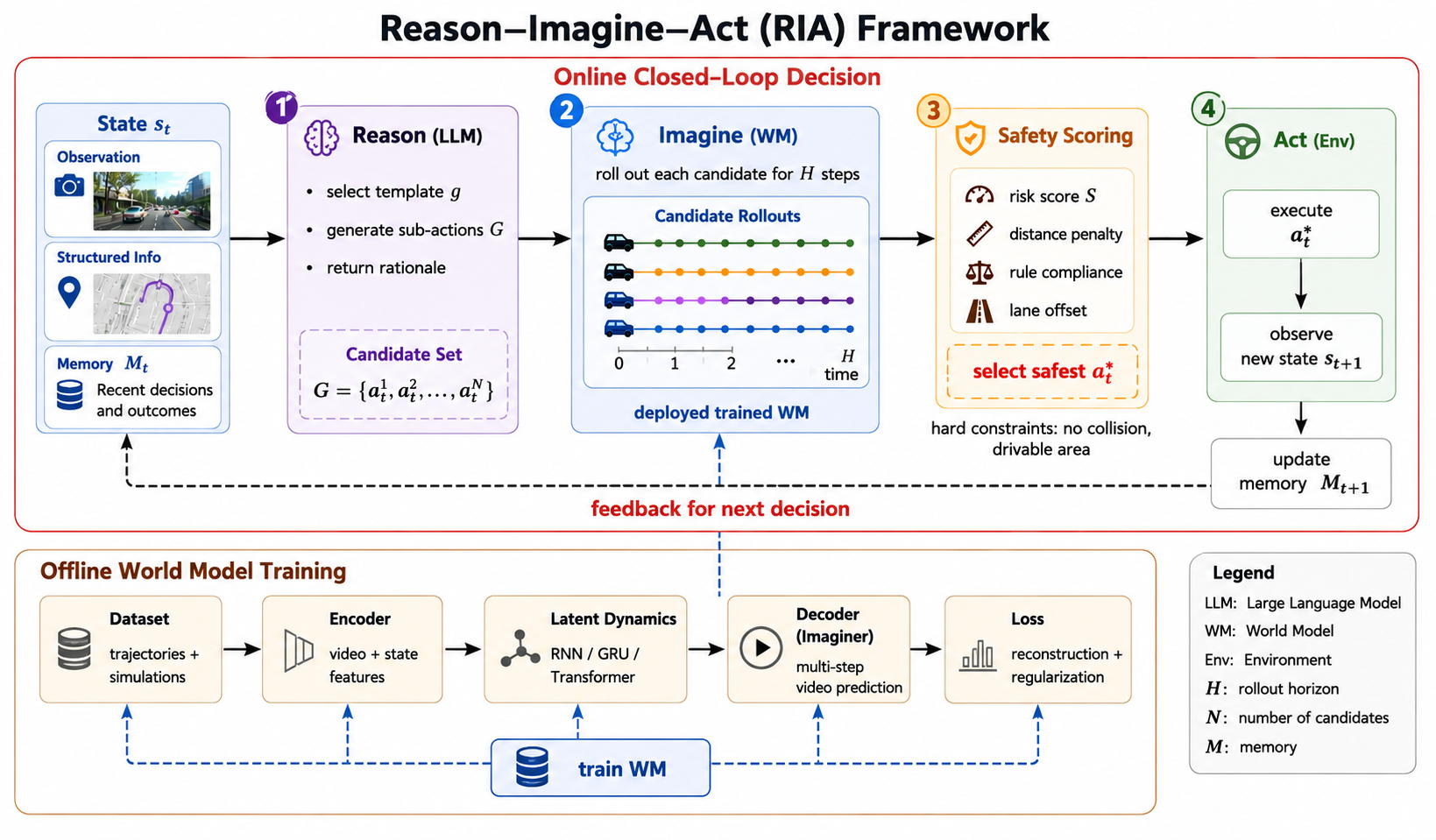}
    \caption{Reason--Imagine--Act (RIA) closed-loop framework. The LLM proposes an action template; the WM performs
    short-horizon rollouts over candidate sub-actions; a safety scorer selects the safest option and feeds back physical
    verification signals to the LLM.}
    \label{fig:ria_framework}
\end{figure*}

Figure~\ref{fig:ria_framework} shows the complete Reason--Imagine--Act closed-loop pipeline.

\textbf{Notation.}
Let $o_t$ denote raw observations (e.g., ego states, traffic lights, nearby agents), and $s_t$ denote the structured state
used by both the LLM and the WM. The LLM outputs a discrete template $g_t \in \mathcal{G}$ and a set of candidates
$\mathcal{U}(g_t)=\{u_t^{(j)}\}_{j=1}^{J}$ (sub-actions) under this template. The WM predicts a rollout trajectory
$\hat{\tau}_t^{(j)} = \{ \hat{s}_{t+1}^{(j)},\ldots,\hat{s}_{t+H}^{(j)}\}$ for each candidate. A safety scorer returns
a scalar score $S(\hat{\tau}_t^{(j)})$ and chooses $u_t^\star=\arg\max_j S(\hat{\tau}_t^{(j)})$ for execution,
where $j$ is the candidate index, $J$ is the number of candidate sub-actions, and $H$ is the rollout horizon (in control steps).

\subsection{World Model Training}
\label{sec:wm_training}
We train an action-conditioned world model to predict short-horizon environment evolution under ego actions.
Our WM follows a latent dynamics formulation inspired by recurrent state-space models (RSSM) \cite{hafner2019learning}.

\paragraph{State representation.}
At each step, we construct a compact structured state $s_t$ by aggregating:
(i) \textbf{ego} features (position, velocity, yaw, acceleration, lane-center offset, route progress),
(ii) \textbf{road} features (lane topology, route direction, speed limit, traffic light state,
distance to stop line, and distance to lane boundaries),
and (iii) \textbf{social} features (relative pose/velocity of surrounding vehicles and pedestrians).
The action input $a_t$ corresponds to the executed low-level control signal (or a sub-action parameterization)
used during data collection.
These geometric quantities make rule and road violations (e.g., lane-boundary crossing, off-road, and red-light related crossing risk)
directly computable from predicted states.

\paragraph{Social attention encoding.}
To model multi-agent coupling around the ego vehicle, we use an attention-based social encoder over nearby agents.
Given the ego token $x_e$ and neighbor tokens $\{x_i\}_{i=1}^{N}$ at step $t$, we compute
\begin{equation}
\alpha_i=\mathrm{softmax}_i\!\left(\frac{(W_qx_e)^\top(W_kx_i)}{\sqrt{d}}\right),\quad
c_t=\sum_{i=1}^{N}\alpha_i\,W_vx_i,
\end{equation}
where $c_t$ is the social context feature.
This design allows the WM to prioritize interaction risk instead of raw distance, so high-closing-speed cut-ins can dominate attention even if they are not the nearest objects.

\paragraph{Model architecture.}
The WM maintains a latent belief state $h_t$ updated recurrently:
\begin{equation}
h_t = f_{\text{dyn}}(h_{t-1}, z_{t-1}, a_{t-1}),
\end{equation}
where $z_t$ is the stochastic latent variable inferred from the current observation/state.
Here $f_{\text{dyn}}(\cdot)$ denotes the latent transition function.
In implementation, the encoder follows a spatial-temporal factorization.
The spatial branch computes social context $c_t$ from ego-neighbor interaction tokens, and the temporal branch applies a GRU over the fused sequence
$x_t=[s_t;c_t]$ to encode second-order trends (e.g., acceleration and yaw-rate evolution):
\begin{equation}
\tilde{h}_t=\mathrm{GRU}(x_t,\tilde{h}_{t-1}).
\end{equation}
The fused latent feature is used to infer $q(z_t|s_t,h_t)$, and the decoder predicts residual dynamics
\begin{equation}
\hat{s}_{t+1}=s_t+g_{\text{dec}}(h_t,z_t,a_t),
\end{equation}
which is numerically more stable than direct absolute-state regression in closed-loop rollouts.
We use action-conditioning to model trajectory bifurcation under different candidate controls.

\paragraph{Data collection.}
We collect trajectories by combining an expert policy (e.g., CARLA Traffic Manager) with controlled perturbations
to cover both nominal and near-failure regimes (e.g., close cut-ins, hard braking, lane boundary violations).
Specifically, we use expert-guided noise exploration with $\epsilon=0.2$ to increase coverage of near-collision and off-road precursor states.
To strengthen boundary learning, we additionally keep short critical windows immediately before failure events
(collision, off-road, and lane-boundary crossing) in the training set.
The dataset consists of tuples $(s_t,a_t,s_{t+1})$ with fixed control frequency.
In each tuple, $s_t$ is the current structured state, $a_t$ is the executed action at step $t$, and $s_{t+1}$ is the next-step state.

\paragraph{Training objective.}
We optimize the WM with a multi-step prediction loss over horizon $H$:
\begin{equation}
\mathcal{L}_{\text{WM}} = \frac{1}{H}\sum_{k=1}^{H} \gamma^k \left\| \hat{s}_{t+k} - s_{t+k} \right\|_2^2
\;+\; \lambda\,\mathcal{L}_{\text{latent}},
\end{equation}
where $\gamma\in(0,1]$ discounts long-horizon errors and $\mathcal{L}_{\text{latent}}$ denotes standard latent regularization
(e.g., KL terms). Here $k$ is the rollout offset index and $\lambda$ balances prediction accuracy and latent regularization.
We train the WM until short-horizon forecasts are stable enough to support online verification.

\paragraph{Design choices for closed-loop verification.}
Our WM is explicitly configured for online safety screening instead of long-horizon video synthesis.
First, compact structured states make key safety predicates directly measurable from rollouts (lane departure, front-gap collapse, and stop-line risk), without additional perception post-processing.
Second, social attention and temporal recurrence are decoupled: the former selects interaction relevance in the current scene, and the latter tracks motion continuity across time.
Third, residual action-conditioned decoding preserves control sensitivity, so different candidate sub-actions from the same state produce distinguishable counterfactual futures.
Fourth, near-failure-window sampling improves transition modeling around safety boundaries, which is exactly where closed-loop ranking decisions are most sensitive.
Fifth, we use the same state variables for WM rollout scoring and TM control adaptation, keeping the verification semantics consistent from prediction to execution.

\paragraph{From prediction fidelity to decision utility.}
Our objective in closed-loop driving is not long-horizon pixel-level realism, but reliable \emph{candidate ordering} within a short reaction window.
At each decision step, the WM is used to rank a small action set $\mathcal{U}(g_t)$ by predicted safety.
This requires three properties: (i) local geometric accuracy around safety boundaries, (ii) consistent action sensitivity so rollout branches do not collapse, and (iii) semantic consistency between prediction variables and scoring variables.
The above architectural choices are designed around this ranking objective, which is why we prioritize 1\,s rollout fidelity and boundary-transition coverage in training.

\subsection{Agent Design and WM-Augmented Decision Loop}
\label{sec:agent_loop}
Our agent builds upon the modular agent design of MADA \cite{yang2024driving} and replaces direct rule-based arbitration
with an LLM-driven template selection plus WM-based short-horizon verification. We use the DeepSeek API as the LLM backend
\cite{noauthor_your_nodate}.
Crucially, verification outcomes are fed back to the LLM as persistent physical feedback, enabling iterative correction across
time rather than one-shot candidate ranking.

\subsubsection{LLM inputs}
The LLM prompt is structured into three parts:
\begin{itemize}
    \item \textbf{Current State \& Environment.} Ego speed, speed limit, current/next road direction, traffic light state,
    leading vehicles and pedestrians, lane-change permissions, and any detected constraints.
    \item \textbf{Previous Driving Behaviors.} A short memory buffer containing recent decisions, executed sub-actions,
    and key verification outcomes to reduce oscillations (e.g., repeated lane-change toggling).
    \item \textbf{Driver Prompt (System Instruction).} A strict schema that constrains output format and action space,
    requiring structured JSON responses (decision + rationale + candidate set).
\end{itemize}

\subsubsection{Discrete action templates}
The LLM selects one template from a discrete set $\mathcal{G}$:
\begin{itemize}
    \item ``stop'': emergency stop.
    \item ``speed up'' / ``speed down'' / ``maintain speed'': adjust target speed while tracking route.
    \item ``lane changing left'' / ``lane changing right'': switch target lane and re-generate a local path.
    \item ``normal behavior'': continue lane-keeping navigation under traffic rules.
\end{itemize}

Each template induces a small candidate set of \emph{sub-actions} $\mathcal{U}(g_t)$, e.g.,
target speed levels, braking intensity profiles, steering bias parameters, or lane-change aggressiveness settings.
This discretization keeps the reasoning space compact while allowing physically grounded refinement.
In implementation, each template maps to a small predefined intensity group:
speed-up/speed-down/stop use soft-base-hard triplets,
maintain-speed uses minus-base-plus, lane-change templates use gentle-base-fast,
and normal-behavior uses a conservative four-level set (maintain, decelerate, hard decelerate, stop).
If a template has no predefined group, refinement is skipped and the original template action is executed.

\subsubsection{WM-based short-horizon evaluation (Imagine)}
After the LLM chooses $g_t$, the WM evaluates the candidate sub-actions:
\begin{enumerate}
    \item \textbf{Feature extraction.} Build $s_t$ from real-time observations (ego + road + social).
    \item \textbf{Belief update.} Update WM hidden state using the last executed action and the newly observed state.
    \item \textbf{Rollout.} For each candidate $u_t^{(j)}\in\mathcal{U}(g_t)$, roll out $H$ steps to obtain $\hat{\tau}_t^{(j)}$.
    \item \textbf{Safety scoring.} Compute $S(\hat{\tau}_t^{(j)})$ (higher is safer) and select $u_t^\star$.
    \item \textbf{Fallback rules.} Apply hard constraints as pre-rollout short-circuit and post-scoring override (e.g., immediate brake on imminent collision).
\end{enumerate}

\paragraph{Safety scoring.}
In implementation, we first compute a rollout penalty cost
\begin{equation}
C(\hat{\tau}_t^{(j)}) = p_{\mathrm{crit}} + p_{\mathrm{dist}} + p_{\mathrm{lane}},
\end{equation}
where
\begin{equation}
p_{\mathrm{crit}} = 100\cdot \mathbf{1}[d_{\min}<3.0],\quad
p_{\mathrm{dist}}=\max(0,8.0-d_{\min})^2,
\end{equation}
and $d_{\min}$ is the minimum predicted distance to surrounding agents along rollout $\hat{\tau}_t^{(j)}$.
Let $l_{\max}$ be the maximum lateral offset in the rollout, and let $b_{\mathrm{lc}}\in\{0,1\}$ indicate
whether the template is a lane-change action. The lane term is
\begin{equation}
p_{\mathrm{lane}}=b_{\mathrm{lc}}\,p_{\mathrm{lane}}^{\mathrm{lc}}
+(1-b_{\mathrm{lc}})\,p_{\mathrm{lane}}^{\mathrm{nlc}},
\end{equation}
where superscripts $\mathrm{lc}$ and $\mathrm{nlc}$ denote lane-change and non-lane-change cases, respectively.
\begin{equation}
p_{\mathrm{lane}}^{\mathrm{lc}}=2.0\max(0,l_{\max}-2.8),
\end{equation}
\begin{equation}
p_{\mathrm{lane}}^{\mathrm{nlc}}=8.0\max(0,l_{\max}-1.5).
\end{equation}
We select the candidate with minimum cost and use an equivalent safety score definition $S(\hat{\tau})=-C(\hat{\tau})$ in the algorithmic description.
Hard constraints are applied in two stages: a pre-rollout short-circuit for imminent front-collision risk,
and a post-scoring override if all rollout candidates violate critical thresholds.
If there is only one candidate, it is used directly; if rollout fails/returns empty, we fall back to the middle-intensity candidate when available.

\subsubsection{TM-based low-level execution}
Our low-level controller is CARLA Traffic Manager (TM). The RIA module does not directly output
continuous throttle/brake/steer commands; instead, it outputs high-level templates and sub-actions that are mapped
to TM behavior parameters and route-following options (e.g., target speed adjustment, lane-change intent,
and conservative vs. aggressive passing preference). TM then produces executable control commands at each simulator step.
For safety-critical situations, our hard-rule fallback (e.g., emergency braking) can override normal TM behavior.
Sub-actions are mapped to calibrated intensity parameters, e.g., speed-up increments
($0.25/0.5/1.0$), speed-down decrements ($0.8/1.5/2.5$), maintain-speed offsets
($-1/0/+1$ km/h around recent speed), and lane-change speed bias ($-3/0/+3$ km/h).
The soft-stop variant additionally caps braking authority ($\mathrm{brake}\le 0.4$).
TM also consumes four WM hints (front-gap distance, front time-to-collision, normalized lane offset,
and static-obstacle-ahead distance) to form a risk score
\begin{equation}
r=r_{\mathrm{gap}}+r_{\mathrm{ttc}}+r_{\mathrm{static}}+r_{\mathrm{lane}},
\end{equation}
with bucketized increments:
front gap below 3/6/10\,m adds 3/2/1.
TTC below 1.5/2.5/4.0\,s adds 3/2/1.
Static-obstacle distance below 4/8\,m adds 2/1.
Lane-offset norm above 1.1/1.3 adds 1/2.
TM mode is then selected by $r$.
If $r\ge5$, we use high-risk mode (follow distance 5.0\,m, speed reduction at least 30\%, random lane change disabled).
If $3\le r<5$, we use medium-risk mode (follow distance 3.0\,m, speed reduction at least 15\%, restricted lane change).
If $r<3$, we keep the low-risk baseline (follow distance 1.8\,m with template-conditioned speed/lane-change settings).

\subsubsection{Closed-loop execution and feedback (Act)}
The selected sub-action $u_t^\star$ is passed to the TM backend for low-level execution.
We log the executed action and the WM verification summary into the memory buffer, and provide a compact physical feedback
message to the LLM at the next step (e.g., ``predicted side-collision under candidate lane-change; chose mild brake'').
This completes the RIA loop and enables iterative self-correction over time.
In Algorithm~\ref{alg:ria}, $\phi(\cdot)$ maps observations to structured states, $m_t$ is the behavior memory at step $t$,
$r_t$ is the LLM rationale output, $S^{(j)}$ is the score of candidate $u_t^{(j)}$, and $e_t$ is the execution summary.

\begin{algorithm}[t]
\caption{Reason--Imagine--Act (RIA) Decision Loop}
\label{alg:ria}
\begin{algorithmic}[1]
\Require Observation $o_t$, memory $m_{t-1}$, WM hidden state $h_{t-1}$
\State Build structured state $s_t \leftarrow \phi(o_t)$
\State \textbf{Reason:} $(g_t,\mathcal{U}(g_t),r_t)\leftarrow \text{LLM}(s_t,m_{t-1})$
\If{$\text{HardRisk}(s_t)$}
    \State $u_t^\star \leftarrow \text{SafeFallback}(g_t)$; execute $u_t^\star$, obtain $e_t$
    \State $q_t \leftarrow \text{hard\_trigger}$
    \State $m_t \leftarrow \text{UpdateMem}(m_{t-1}, g_t, u_t^\star, e_t, q_t)$
    \State \Return executed action $u_t^\star$
\EndIf
\State Update belief: $h_t \leftarrow \text{WMUpdate}(h_{t-1}, s_t, a_{t-1})$
\ForAll{$u_t^{(j)}\in \mathcal{U}(g_t)$}
    \State \textbf{Imagine:} $\hat{\tau}_t^{(j)} \leftarrow \text{Rollout}(\text{WM}, h_t, s_t, u_t^{(j)}, H)$
    \State Score: $S^{(j)} \leftarrow S(\hat{\tau}_t^{(j)})$
\EndFor
\State $u_t^\star \leftarrow \arg\max_j S^{(j)}$
\State Apply post-scoring hard override when critical constraints are violated
\State \textbf{Act:} execute $u_t^\star$, obtain execution summary $e_t$
\State $q_t \leftarrow \max_j S^{(j)}$
\State $m_t \leftarrow \text{UpdateMem}(m_{t-1}, g_t, u_t^\star, e_t, q_t)$
\State \Return executed action $u_t^\star$
\end{algorithmic}
\end{algorithm}

%% file: sections/experiments.tex
\section{Experiments}
\label{sec:experiments}

\subsection{Experimental Setup}
\label{sec:exp_setup}
\paragraph{Simulator.}
We evaluate in CARLA~0.9.15~\cite{dosovitskiy2017carla} with synchronous stepping and a fixed controller update frequency.
All methods share the same perception interface (privileged simulator state).
Each baseline keeps its native low-level control implementation, while our method uses the TM-based backend described in Sec.~\ref{sec:agent_loop}.

\paragraph{Tasks: random point-goal navigation.}
We consider a randomized point-goal navigation task: each episode samples a start and a goal waypoint,
and the agent must reach the goal under realistic traffic while obeying traffic rules.
We run \textbf{1000} episodes across randomly selected maps with an identical route sampling strategy
and identical traffic generation policy for all compared methods.

\paragraph{Maps.}
We evaluate on multiple CARLA towns and randomly sample map-route combinations during evaluation.

\paragraph{Traffic generation.}
We use CARLA Traffic Manager (TM) to spawn background traffic with identical parameters across methods:
number of vehicles $N_v=30$, number of pedestrians $N_p=8$, global speed difference, minimum following distance,
and traffic light compliance. We also fix weather presets to reduce variance, unless explicitly studying
generalization.

\subsection{World Model Prediction Quality}
\label{sec:wm_quality}
As a prerequisite for closed-loop comparison, we first verify whether the WM is accurate enough for online
candidate screening. We evaluate 1\,s prediction quality on a held-out dataset built with the same
state/action interface used during deployment.
Table~\ref{tab:wm_quality} reports displacement, heading, and speed errors of WM rollouts.
These values indicate that short-horizon rollout errors remain in a regime where relative candidate ranking is stable:
the model preserves lane-level geometry and heading evolution while maintaining usable velocity trends for TTC/front-gap estimation.
This is sufficient for the \emph{Imagine} stage, whose goal is online risk ordering among a small candidate set, rather than exact long-horizon trajectory reconstruction.
For context, Table~\ref{tab:wm_ref_msma} lists reported open-loop trajectory metrics from a recent CARLA-based study \cite{chen2024msma}.

\begin{table}[!htbp]
\centering
\caption{1\,s world-model prediction quality on held-out data.}
\label{tab:wm_quality}
\setlength{\tabcolsep}{4.0pt}
\renewcommand{\arraystretch}{1.08}
\begin{tabular}{lc}
\toprule
Metric & Value \\
\midrule
ADE @ 1\,s & 0.34\,m \\
FDE @ 1\,s & 0.85\,m \\
Yaw RMSE @ 1\,s & 1.2$^\circ$--2.5$^\circ$ \\
Velocity RMSE @ 1\,s & 0.3--0.6\,m/s \\
\bottomrule
\end{tabular}
\end{table}

\begin{table}[!htbp]
\centering
\caption{Reference metrics from \cite{chen2024msma} (open-loop CARLA trajectory prediction).}
\label{tab:wm_ref_msma}
\scriptsize
\setlength{\tabcolsep}{3.6pt}
\renewcommand{\arraystretch}{1.05}
\begin{tabular}{lccc}
\toprule
Setting in \cite{chen2024msma} & ADE (m) & FDE (m) & MR \\
\midrule
MSMA, MPR=0.0 & 0.62 & 1.48 & 0.23 \\
MSMA, MPR=0.8 & 0.56 & 1.33 & 0.20 \\
HiVT baseline (ADE only) & 0.66 & -- & -- \\
\bottomrule
\end{tabular}
\end{table}
{\footnotesize Note: this external table is for scale reference only. Protocols differ in objective (open-loop trajectory prediction vs. action-conditioned rollout verification), horizon, evaluated agent set, and metric aggregation.}

\begin{table*}[!t]
\centering
\caption{Closed-loop point-goal navigation results on 1000 episodes with safety/comfort indicators.}
\label{tab:main_results}
\setlength{\tabcolsep}{4.2pt}
\renewcommand{\arraystretch}{1.12}
\begin{tabular}{lccccccccc}
\toprule
Method & RC (\%)$\uparrow$ & AR (\%)$\uparrow$ & ColR (\%)$\downarrow$ & RedL$\downarrow$ & Stop$\downarrow$ & OffR$\downarrow$ & HB$\downarrow$ & HJ$\downarrow$ & MAJ$\downarrow$ \\
\midrule
LLM w/o WM verification & 61.21 & 30.20 & 0.40 & 0.03 & \textbf{0.01} & 0.28 & 207.41 & 87.85 & 97.43 \\
CARLA TM & 46.47 & 13.40 & 16.00 & 0.75 & 0.12 & \textbf{0.00} & 33.80 & 181.87 & 72.80 \\
MADA baseline policy & 21.88 & 22.40 & 53.00 & 0.30 & 0.19 & 1.71 & 289.40 & 158.62 & 51.30 \\
\textbf{LLM+WM (Ours)} & \textbf{80.05} & \textbf{51.10} & \textbf{0.20} & \textbf{0.01} & 0.09 & 0.19 & \textbf{24.58} & \textbf{54.77} & \textbf{31.21} \\
\bottomrule
\end{tabular}
\vspace{0.25em}

\footnotesize RedL: red-light violations/route; Stop: stop-sign violations/route; OffR: offroad events/route;
HB: hard-brake events/route; HJ: high-jerk events/route; MAJ: mean absolute jerk.
\end{table*}

\subsection{Compared Methods}
\label{sec:baselines}
We evaluate under an inference-time, no-parameter-update deployment constraint.
All compared agents run with the same CARLA closed-loop interface and evaluation protocol,
without additional policy/value training or fine-tuning in our setup.
Under this constraint set, MADA is used as the primary external baseline, and TM/LLM variants
are included as internal baselines to isolate the contribution of WM verification.
We compare the following methods.

\paragraph{RIA (ours, with WM verification and feedback).}
Our \textbf{Reason--Imagine--Act} agent uses an LLM to select a discrete action template and candidate sub-actions.
The world model performs short-horizon rollout verification and the safest candidate is executed through TM.

\paragraph{LLM w/o WM verification.}
This variant keeps the same LLM and action space but removes world-model verification.
Without WM risk cues, the TM execution layer tends to select relatively conservative actions.

\paragraph{CARLA TM.}
The native Traffic Manager policy in CARLA \cite{dosovitskiy2017carla} is used as a non-LLM reference baseline.

\paragraph{MADA baseline policy.}
We include the MADA policy \cite{yang2024driving} as a strong LLM-driven baseline under the same
deployment constraint and interface.
The main quantitative comparison is reported in Sec.~\ref{sec:main_results}
(Table~\ref{tab:main_results}).

\subsection{Metrics}
\label{sec:metrics}
Following the finalized evaluation protocol in this work, aligned with CARLA Leaderboard-style evaluation
and prior CARLA closed-loop studies \cite{noauthor_evaluation_nodate,chen2020learning,li2024think2drive}, we report:
\begin{itemize}
    \item \textbf{Average Route Completion (RC, \%)}: average percentage of route distance completed.
    \item \textbf{Arrival Rate (AR, \%)}: percentage of episodes that successfully reach the goal.
    \item \textbf{Collision Rate (ColR, \%)}: percentage of episodes with at least one collision.
    \item \textbf{Red-light Violations per Route (RedL)}: average number of red-light violations per route.
    \item \textbf{Stop-sign Violations per Route (Stop)}: average number of stop-sign violations per route.
    \item \textbf{Offroad Events per Route (OffR)}: average number of offroad events per route.
    \item \textbf{Hard-brake Events per Route (HB)}: average number of hard-brake events per route.
    \item \textbf{High-jerk Events per Route (HJ)}: average number of high-jerk events per route.
    \item \textbf{Mean Absolute Jerk (MAJ)}: average absolute jerk magnitude over executed trajectories.
\end{itemize}
The comfort-related jerk indicators are reported as supplementary behavior quality signals, which are common in trajectory/behavior analysis settings \cite{chen2023follownet}.

\subsection{Implementation Details}
\label{sec:implementation}
We use the DeepSeek~V3.2 API~\cite{noauthor_your_nodate} as the reasoning module and keep the same template/sub-action
space for all LLM-based variants. The deployed configuration follows Sec.~\ref{sec:agent_loop}: fixed JSON-structured
prompting, template-conditioned candidate generation, 1\,s WM rollout screening, and hard-rule fallback in safety-critical cases.
Here $J$ is the number of candidates per template and $H$ is the rollout horizon.
\textbf{Code Availability.} Code and evaluation scripts are available at \url{https://github.com/pku-smart-city/source_code/tree/main/RIA}.

\subsection{Experimental Findings}
\label{sec:main_results}
Table~\ref{tab:main_results} reports the closed-loop results. We summarize the key findings as follows.

\paragraph{Finding 1: WM verification is the primary source of gain.}
Compared with \textbf{LLM w/o WM verification}, adding WM rollout screening improves RC from 61.21\% to 80.05\%,
AR from 30.20\% to 51.10\%, and reduces ColR from 0.40\% to 0.20\%.
Behavior-level metrics also improve substantially: OffR drops from 0.28 to 0.19, HB from 207.41 to 24.58,
HJ from 87.85 to 54.77, and MAJ from 97.43 to 31.21.

\paragraph{Finding 2: RIA gives the strongest overall closed-loop task performance.}
Against \textbf{CARLA TM} and \textbf{MADA}, RIA achieves the best RC/ColR profile
(RC: 80.05 vs. 46.47/21.88, ColR: 0.20 vs. 16.00/53.00) and the best AR (51.10 vs. 13.40/22.40),
showing clear gains in route completion and safety outcomes.

\paragraph{Finding 3: Most safety/comfort indicators improve, with explicit trade-offs.}
RIA is lower than both external baselines on RedL, Stop, HB, HJ, and MAJ.
Relative to MADA, RIA is also much lower on OffR (0.19 vs. 1.71).
Relative to TM, the main trade-off is OffR (0.19 vs. 0.00), while RIA remains clearly better on RC/AR/ColR.
Taken together, this indicates a stronger overall balance under the same closed-loop interface.

%% file: sections/conclusion.tex
\section{Conclusion}
\label{sec:conclusion}
We presented \textbf{Reason--Imagine--Act}, a closed-loop autonomous driving framework that bridges
high-level LLM reasoning with physically grounded decision making via an action-conditioned world model.
By evaluating candidate sub-actions through short-horizon rollouts and feeding back physical verification
signals to the LLM, our system mitigates physically infeasible or unsafe decisions that can arise from
purely semantic policies.

Extensive experiments in CARLA~0.9.15 on randomized point-goal navigation across multiple towns show that
our complete RIA configuration achieves higher route completion and success rates while substantially reducing collisions
and infractions compared to an LLM-only baseline, with consistent gains over strong training-free baselines
under the same closed-loop interface.